\documentclass[10pt, twocolumn]{article}

\usepackage[a4paper, total={7in, 9.77in}]{geometry}

\usepackage{newtxtext}
\usepackage{amsmath,amsfonts,amssymb}
\usepackage{algorithmic, algorithm}
\usepackage{array, textcomp, stfloats, verbatim}
\usepackage{graphicx, cite, booktabs, cases, contour, placeins}

\usepackage{indentfirst}
\usepackage{titlesec}
\titlespacing{\section}{0pt}{5pt plus 1pt minus 1pt}{4pt plus 1pt minus 1pt}
\titlespacing{\subsection}{0pt}{4pt plus 1pt minus 1pt}{3pt plus 1pt minus 1pt}
\titlespacing{\subsubsection}{0pt}{3pt plus 1pt minus 1pt}{2pt plus 1pt minus 1pt}

\makeatletter
\def\@seccntformat#1{\csname the#1\endcsname\quad}
\makeatother

\usepackage{fancyhdr}
\title{Decoupled Functional Evaluation of Autonomous Driving Models via Feature Map Quality Scoring}
\author{
    Ludan Zhang\textsuperscript{1,2,3} \quad
    Sihan Wang\textsuperscript{1,2,3} \quad
    Yuqi Dai\textsuperscript{1,2} \quad 
    Shuofei Qiao\textsuperscript{4} \quad
    Qinyue Luo\textsuperscript{5} \quad
    Lei He\textsuperscript{1,2,} \thanks{Corresponding author: helei2023@tsinghua.edu.cn \\
\textsuperscript{1}School of Vehicle and Mobility, Tsinghua University, Beijing 100084, China.\\
\textsuperscript{2}State Key Laboratory of Intelligent Green Vehicle and Mobility, Tsinghua University, Beijing 100084, China.\\
\textsuperscript{3}School of Mathematical Sciences, Nankai University, Tianjin, China.\\
\textsuperscript{4}College of Computer Science and Technology, Zhejiang University, Hangzhou, China.\\
\textsuperscript{5}SAIC-GM-Wuling Automobile Co., Ltd. Liuzhou 545007, PR China.} \quad
}
\date{}
\begin{document}

\maketitle

\begin{abstract}
End-to-end models are emerging as the mainstream in autonomous driving perception and planning. However, the lack of explicit supervision signals for intermediate functional modules leads to opaque operational mechanisms and limited interpretability, making it challenging for traditional methods to independently evaluate and train these modules. Pioneering in the issue, this study builds upon the feature map-truth representation similarity-based evaluation framework and proposes an independent evaluation method based on Feature Map Convergence Score (FMCS). A Dual-Granularity Dynamic Weighted Scoring System (DG-DWSS) is constructed, formulating a unified quantitative metric—Feature Map Quality Score—to enable comprehensive evaluation of the quality of feature maps generated by functional modules. A CLIP-based Feature Map Quality Evaluation Network (CLIP-FMQE-Net) is further developed, combining feature-truth encoders and quality score prediction heads to enable real-time quality analysis of feature maps generated by functional modules. Experimental results on the NuScenes dataset demonstrate that integrating our evaluation module into the training improves 3D object detection performance, achieving a 3.89 percent gain in NDS. These results verify the effectiveness of our method in enhancing feature representation quality and overall model performance.
\end{abstract}

\AtBeginEnvironment{tabular}{\footnotesize}
\setlength{\parindent}{2em} 
\begin{figure*}[t]
  \centering
  \includegraphics[width=\linewidth]{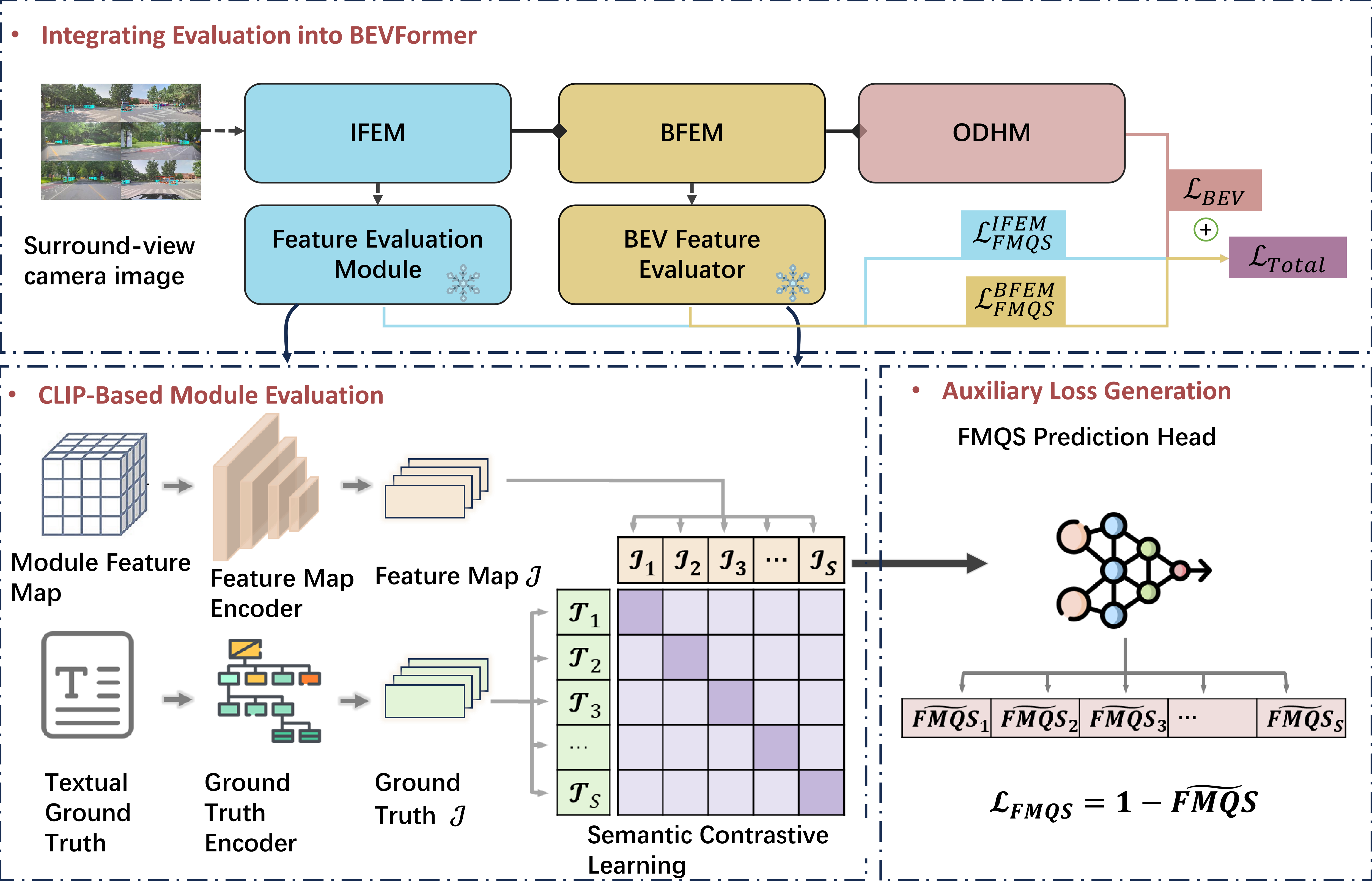}
  \caption{\textbf{Independent Evaluation Method for Complex Modules Based on FMQS. } CLIP-FMQE-Net enables quality assessment of feature maps by aligning them with structured ground truth in a shared semantic space and predicting scores via Transformer-based decoding.}
  \label{fig1}
\end{figure*}
\section{Introduction}

End-to-end deep learning algorithms have been widely utilized in autonomous driving perception systems due to their powerful feature learning capabilities and streamlined training processes.  However, their black-box nature leads to a lack of explicit supervision for intermediate features and obstructs the direct association between internal representations and final task performance. Consequently, traditional evaluation methods are insufficient for effectively optimizing functional modules. 

Multi-Module Learning (MML) \cite{dai2024hierarchical} provides a promising approach by dividing perception algorithms into multiple specialized functional modules. This enables independent training and flexible deployment in diverse scenarios, and helps improve interpretability and reusability. However, it still lacks unified and quantitative criteria for assessing the training maturity of each module. Feature maps play a central role in transferring information between modules. Evaluating the performance of each module based on the quality of its output feature maps has become a key challenge. Addressing this issue is critical for improving model development efficiency and ensuring system-level safety.

Therefore, it is essential to establish a systematic evaluation mechanism for module training maturity, which will improve the interpretability and optimization efficiency in MML. FMCE-Net \cite{zhang2024feature} proposes an initial approach for evaluating functional modules by scoring feature maps through an analysis of loss values on the image classification task. While effective for models with a single functional module, this method lacks the capacity to independently evaluate multiple cascaded functional modules. 

To address this limitation, a recent study \cite{zhang2024unveiling} proposed a feature map-truth representation similarity-based evaluation method. By leveraging a two-stage aligned autoencoders and large language models, both feature maps and ground truth information are projected into a unified semantic representation space, where the Representation Similarity Score (RSS) is employed to quantify the module training maturity. While the method reveals a significant positive correlation between RSS and perception task performance metrics on BEVFormer, it is not yet integrated into the model training process and therefore cannot directly contribute to optimization.

To address the challenge of integrating functional module evaluation into complex model training and optimization, we propose a dynamic evaluation mechanism based on Feature Map Quality Score (FMQS). Specifically, we design a dynamic weighted scoring system for dual-granularity maturity analysis, tailored for MML training. This mechanism assesses module training maturity from two perspectives: the global model level and the local feature map level. It integrates feature map semantic consistency and task performance feedback to establish a multidimensional quantitative assessment model for module training maturity.

To implement this approach, we develop a CLIP-based Feature Map Quality Evaluation Network (CLIP-FMQE-Net), which saves the output feature maps and ground truth of each functional module during MML. These are then served as inputs to CLIP's image and text encoders, respectively. By aligning feature maps with ground truth, the network leverages a Transformer-based decoder to accurately predict the FMQS, thereby enabling a closed-loop evaluation process. Finally, the proposed evaluation module is effectively integrated into the training process of 3D object detection tasks, allowing for adaptive guidance and continuous optimization.

Our main contributions are as follows:
\begin{itemize}
    \item This paper proposes FMQS and a dual-granularity dynamic weighted scoring system for evaluating functional module training maturity.
    \item This work designs CLIP-FMQE-Net for closed-loop evaluation of functional modules via feature-truth alignment and regression.
    \item We validate FMQS on BEVFormer, showing improved training efficiency and optimization performance.
\end{itemize}
\section{Related Work}
\subsection{Modular Networks}
In deep learning, neural architectures typically adopt a modular construction approach, consisting of a backbone, neck, head, and multiple functional modules. Each module contains one or more interchangeable algorithmic components. For instance, the DETR network \cite{zhu2020deformable} consists of a backbone, encoder, decoder, and prediction head, with the backbone made up of algorithms such as CNNs. The MNN network \cite{andreas2016neural} decomposes complex tasks into subtasks handled by independent neural modules. PathNet \cite{fernando2017pathnet} is constructed from multilayered modules and employs path evolution to embed agents for identifying transferable components of the network. Dai et al. \cite{dai2024hierarchical} proposes Multi-Module Learning for a hierarchical and decoupled training paradigm for BEV perception. In \cite{zhang2024feature}, an independent method was introduced to evaluate the convergence of functional module outputs in the context of image classification.
\subsection{Contrastive Language-Image Pre-training}
Cross-modal pre-training models have attracted increasing attention in autonomous driving perception tasks. CLIP (Contrastive Language–Image Pretraining) \cite{radford2021learning} maps images and text into a unified semantic space via contrastive learning, demonstrating strong general visual-semantic understanding capabilities. Recent studies, such as CLIP-BEVFormer \cite{pan2024clip} and BEVCLIP \cite{jia2024bev}, incorporate textual information into BEV perception by aligning image-text features. These methods exhibit remarkable robustness and accuracy in complex driving scenarios.
\subsection{Quality Evaluation Method for Feature Maps}
As perception models grow in complexity, the evaluation of feature map quality has gained increasing attention in multi-module architectures. Existing approaches can be broadly categorized into three types. The first category leverages statistical metrics, such as mutual information \cite{maes2002multimodality} and structural similarity \cite{wang2004image}, to assess spatial consistency of feature maps before and after view transformation or alignment. The second category is model-driven, utilizing supervised learning to predict quality labels as an evaluation metric \cite{muenzing2012supervised}, or employing unsupervised methods like autoencoders to assess feature map quality based on reconstruction error \cite{zhang2021anomaly}. The third category employs model interpretability tools—for instance, LIME \cite{ribeiro2016should} and SHAP \cite{wang2021shapley} analyze the contribution of features to model predictions. Grad-CAM \cite{chattopadhay2018grad} visualizes intermediate outputs to support the understanding of feature map quality and internal mechanisms of the model.
\subsection{Independent Module Evaluation in BEVFormer}
For complex models with cascaded multifunctional modules, \cite{zhang2024unveiling} introduced a Representational Similarity Score (RSS) into BEVFormer \cite{li2024bevformer} to independently evaluate module training maturity. Their approach leverages two-stage aligned autoencoders and large language models, feature maps and ground truth information are mapped into a unified semantic representation space, with RSS serving as the evaluation metric. However, this method has not integrated the evaluation metric into the training process, and thus cannot directly guide end-to-end optimization.
\begin{figure*}[t]
  \centering
  \includegraphics[width=\linewidth]{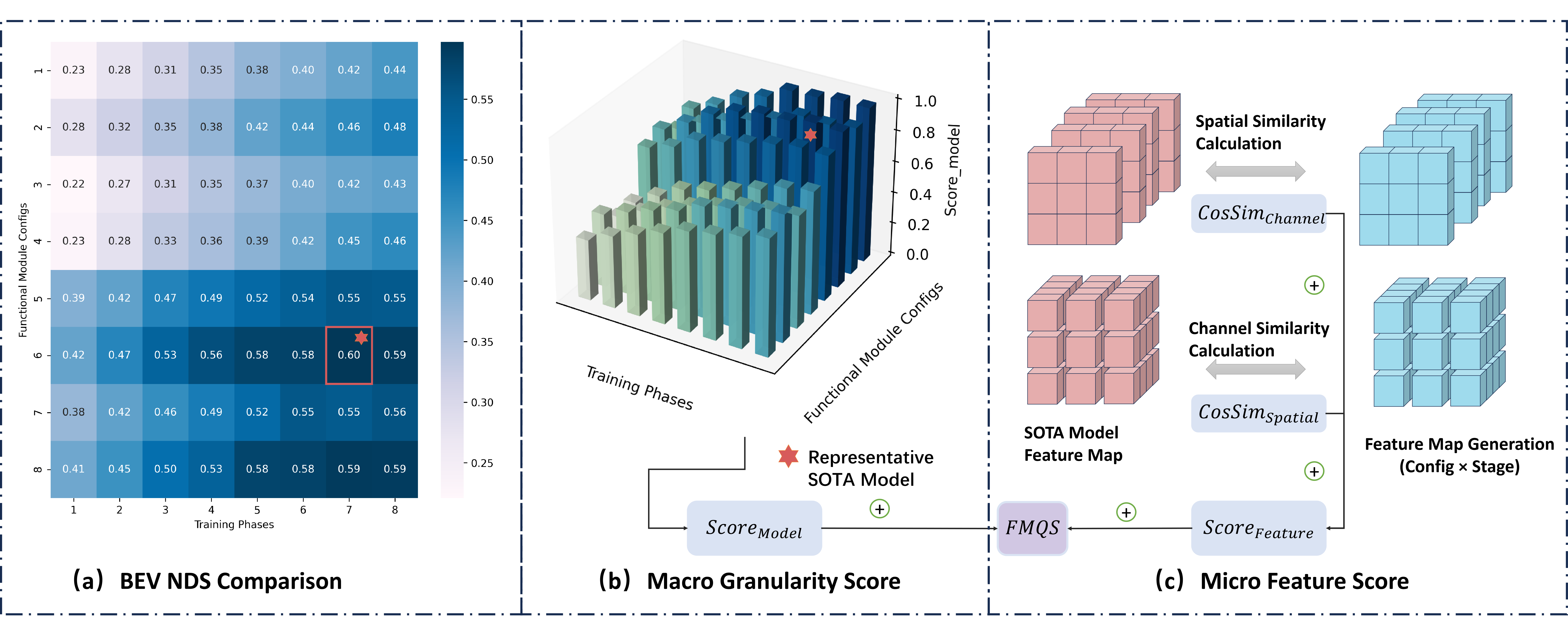}
  \caption{\textbf{Feature Map Quality Score (FMQS) Generation Process.} Macro-level scores are computed via NDS normalization across configurations, while micro-level scores measure feature similarity to the SOTA model using CS-CosSim.}
  \label{fig2}
\end{figure*}
\section{Method}
In end-to-end models, intermediate features often lack explicit supervision signals and are difficult to directly associate with final task metrics, making it challenging for traditional evaluation methods to effectively optimize functional modules. Building upon the feature map–truth representation similarity-based evaluation framework proposed in \cite{zhang2024unveiling}, our work proposes an independent evaluation method based on Feature Map Convergence Score (FMQS). Specifically, we constructe a Dual-Granularity Dynamic Weighted Scoring System (DG-DWSS) and develop a CLIP-based Feature Map Quality Evaluation Network (CLIP-FMQE-Net), which enables effective integration of the evaluation module into the 3D object detection task training process.
\subsection{Feature Map Quality Evaluation System}
To enable independent evaluation of functional modules in complex models based on the Feature Map Quality Score (FMQS), we propose a CLIP-based Feature Map Quality Evaluation Network (CLIP-FMQS-Net), which includes three key components: a feature map encoder, a ground truth text encoder, and an FMQS prediction head. The feature map encoder utilizes a lightweight convolutional network to extract semantic representations from BEVFormer feature maps. The ground truth text encoder transforms structured 2D/3D annotations into text and adopts the CLIP Text Transformer with pretrained weights. By aligning features in a shared semantic space, the ground truth provides effective supervision to improve encoding quality. The FMQS prediction head, built on a Transformer Decoder, leverages self-attention to capture global context for accurate score estimation. The overall framework is shown in Figure~\ref{fig1}.
\subsection{Dual-Granularity Maturity Scoring Mechanism}
To facilitate the training of BEVFormer with multi-module architectures, we establish a unified feature map quality assessment mechanism to evaluate the training maturity of functional modules and utilize it as an auxiliary loss to optimize the model. During MML, we define \(N\) module configurations (e.g., Image Feature Extraction, BEV Representation Generation). At \(M\) key training stages, for each configuration \(C_i\) \( \left( i = 1, 2, \cdots, N \right) \)
, we store the model weights \(S_{ij}\) \( \left( j = 1, 2, \cdots, M \right) \), along with the 3D detection performance metrics (e.g., NDS). This process generates \(N\times M\) combinations of module configuration and training stage, each producing a distinct feature map that reflects the model state under varying training maturity and module composition. Subsequently, the feature map sets under each configuration-stage weight \(S_{ij}\) are obtained during training or centralized inference as:
\begin{equation}
  \begin{aligned}
  \mathcal{F}_{img}^{i,j} &= \{ \mathrm{IFEM}_{ij}(\mathcal{X}_{img}) \mid i = 1,\cdots,N;\ j = 1,\cdots,M \}, \\
  \mathcal{F}_{bev}^{i,j} &= \{ \mathrm{BFEM}_{ij}(\mathcal{F}_{img}^{i,j}) \mid i = 1,\cdots,N;\ j = 1,\cdots,M \}, 
  \end{aligned}
\end{equation}
where \(F_{*}^{i,j}\) denotes the feature map generated at the \(j-th\) training stage under the \(i-th\) configuration.

To fully leverage the advantages of MML, we propose a dual-granularity dynamic weighted scoring system, designed by comparing with the optimal model's features. This mechanism includes:
\subsubsection{Optimal Configuration–Stage Selection}
Since the NuScenes’ NDS metric integrates multiple sub-metrics to assess the overall performance of detection algorithms comprehensively, we adopt NDS as the evaluation criterion and select the configuration–stage pair with the highest NDS among all \(N\times M\) combinations as the SOTA model.
\subsubsection{Macro-Level Granularity Metric}
Let $Score_{model}^{i,j}$ denote the model task granularity score corresponding to the feature map produced by the \(i-th\) configuration at the \(j\-th\) training stage. The current SOTA model is identified as the \(i_{sota}-th\) configuration and the \(j_{sota}-th\) stage yielding the highest NDS, denoted as \(NDS_{sota}\). The model task granularity score for the SOTA combination is defined as:$Score_{model}^{i_{sota},j_{sota}}=1$.
The score for any other combination is computed as the ratio between its NDS and that of the SOTA model:
\begin{equation}
  \begin{aligned}
  Score_{model}^{i,j}=\frac{NDS_{i,j}}{NDS_{sota}}. 
  \end{aligned}
\end{equation}
This produces a model evaluation score matrix $Score_{model}$ of size \(N\times M\), reflecting the model’s performance compared to the SOTA model under different training configurations, with higher NDS indicating more mature and effective module functionality.
\subsubsection{Micro-Level Granularity Metric}
At the feature map level, we introduce the Channel–Spatial Cosine Similarity (CS-CosSim) as a quality metric to quantify the structural similarity between any given feature map and the corresponding feature map from the SOTA model. CS-CosSim jointly considers both the channel and spatial dimensions, making it more effective in preserving the structural information of tensors compared to Cosine similarity. Therefore, we adopt CS-CosSim in this work to assess similarity between feature maps.
The complete procedure consists of three steps:

\begin{enumerate}
\item Let the number of samples in the dataset be denoted by \(S\). Extract \(N\times M\times S\) feature maps from \(N\times M\) configuration–stage combinations, using the SOTA model feature map \( F_{\text{SOTA}} \)  as the reference;

\item Given the \(k\) feature map  \( \mathcal{F} \in \mathbb{R}^{C \times H \times W} \) extracted from the \(i-th\) configuration and the \(j-th\) training stage, we compute its CS-CosSim with the the corresponding SOTA model feature map \(\mathcal{F}_{\text{SOTA}} \) as follows:
\begin{equation}
    \begin{aligned}
        Score_{Feature}^{i,j,k} &= \text{CS-CosSim}(\mathcal{F}^{i,j,k}, \mathcal{F}_{\text{SOTA}}^{i,j,k}) \\
        &= \alpha \cdot CosSim_{Channel}^{i,j,k} \\
        &\quad + (1-\alpha) \cdot CosSim_{Spatial}^{i,j,k}.
    \end{aligned}
\end{equation}

Here, the weighting factor $\alpha$ is set to 0.5 to ensure equal contributions from the channel and spatial dimensions to the overall feature similarity. The channel similarity (Channel-CosSim) $CosSim_{Channel}^{i,j,k}$ and spatial similarity (Spatial-CosSim) $CosSim_{Spatial}^{i,j,k}$ are defined as follows:
\begin{equation}
  \begin{aligned}
   CosSim_{Channel}^{i,j,k} = \frac{1}{C} \sum_{c=1}^C \frac{\mathcal{F}_c^{i,j,k} \cdot \mathcal{F}_{\text{SOTA},c}^{i,j,k}}{\|\mathcal{F}_c^{i,j,k}\| \cdot \|F_{\text{SOTA},c}^{i,j,k}\|}, 
    \end{aligned}
\end{equation}

\begin{equation}
  \begin{aligned}
   CosSim_{Spatial}^{i,j,k} = \frac{1}{H \times W} \sum_{h,w} \frac{\mathcal{F}_{h,w} ^{i,j,k}\cdot \mathcal{F}_{\text{SOTA},h,w}^{i,j,k}}{\|\mathcal{F}_{h,w}^{i,j,k}\| \cdot \|\mathcal{F}_{\text{SOTA},h,w}^{i,j,k}\|}.
     \end{aligned}
\end{equation}
\item Perform a weighted fusion of the model-level score $Score_{model}$ and the feature map similarity $Score_{Feature}$ to obtain the Feature Map Quality Score (FMQS) as follows:
 \begin{equation}
      \begin{aligned}
    \text{\(FMQS\)}^{i,j,k} = w \cdot Score_{model}^{i,j} + (1-w) \cdot Score_{Feature}^{i,j,k},
      \end{aligned}
    \end{equation}
Here, the weighting coefficient \( w \) is set to 0.8.
\end{enumerate}

FMQS provides an efficient, unified, and scalable metric for evaluating the training status of functional modules within multi-module learning frameworks. It facilitates large-scale configuration optimization and enables robust training of BEVFormer-based perception models.The overall FMQS generation process is illustrated in Figure \ref{fig2}.
\subsection{CLIP-FMQE-Net}
To enable quantitative evaluation of feature map quality for functional modules, we propose the CLIP-based Feature Map Quality Evaluation Network(CLIP-FMQE-Net), an independent evaluation model. It consists of two components: a CLIP-based feature map–ground truth encoder and an FMQS prediction head. The encoder aligns feature maps with ground-truth representations in a cross-modal semantic space through contrastive learning, thereby enhancing their semantic representation quality. The prediction head, built on a Transformer architecture, leverages self-attention mechanisms to aggregate semantic information and performs regression to predict FMQS.

The CLIP-based feature map–ground truth encoder consists of two submodules: a ground-truth encoder and a feature map encoder.

The ground truth encoder adopts the CLIP text encoder architecture and leverages its pretrained weights to extract high-quality semantic representation vectors \( T \in \mathbb{R}^{512} \). To ensure semantic consistency across samples, we design ground-truth textual templates based on the nuScenes labeling schema to encode object attribute information \cite{caesar2020nuscenes}. The input is processed through tokenization, embedding, and a multilayer Transformer encoder, ultimately generating a semantic reference for cross-modal alignment. Structured textual descriptions are generated through template filling and then fed into the CLIP text encoder to obtain the semantic vector \( T \in \mathbb{R}^{512} \).

The feature map encoder \(Encoder_{FM}\) adopts a lightweight convolutional neural network comprising four convolutional layers with ReLU activations and pooling layers. It encodes feature map \( F \in \mathbb{R}^{C \times H \times W} \), generated by different functional modules, into semantic representations \( \mathcal{I} \in \mathbb{R}^{512} \) that are aligned with the ground-truth vector. This design facilitates seamless integration into multi-module systems, improving both encoding efficiency and representational capacity. The input–output mapping of the feature map–truth encoder for the IFEM module is defined as:
\begin{equation}
  \begin{aligned}
    \mathcal{T}_{IFEM} = &Encoder_{2DGT}(\mathcal{GT}_{2D}), \\
    \mathcal{I}_{IFEM} = &Encoder_{IFEM}(\mathcal{F}_{img}).
  \end{aligned}
\end{equation}
The input–output mapping of the feature map–truth encoder for the \(BFEM\) module is defined as:
\begin{equation}
  \begin{aligned}
    \mathcal{T}_{BFEM} = &Encoder_{3DGT}(\mathcal{GT}_{3D}), \\
    \mathcal{I}_{BFEM} = &Encoder_{BFEM}(\mathcal{F}_{bev}).
  \end{aligned}
\end{equation}

To achieve effective alignment between the feature maps and the semantic space of ground-truth text, we adopt a symmetric contrastive loss function:
\begin{equation}
  \begin{aligned}
    \mathcal{L}_{\text{cont}} = -\frac{1}{S} \sum_{i=1}^S \Bigg[ 
    &\log \frac{e^{\text{sim}(\mathcal{I}_i, \mathcal{T}_i)/\tau}}{\sum_{j=1}^S e^{\text{sim}(\mathcal{I}_i, \mathcal{T}_j)/\tau}} \\
    +\,&\log \frac{e^{\text{sim}(\mathcal{T}_i, \mathcal{I}_i)/\tau}}{\sum_{j=1}^S e^{\text{sim}(\mathcal{T}_j, \mathcal{I}_i)/\tau}} 
    \Bigg],
  \end{aligned}
\end{equation}
where \( \text{sim}(a,b) = \frac{a \cdot b}{\|a\| \|b\|} \) denotes cosine similarity, and \( \tau \) is a temperature coefficient. This loss encourages high similarity between positive pairs while pushing apart negative pairs, effectively enforcing cross-modal semantic alignment and enhancing semantic expressiveness.

The second component, the FMQS prediction head, regresses the FMQS from the aligned feature representation I. We design a regression network \( \phi: \mathbb{R}^{512} \rightarrow \mathbb{R} \) based on a Transformer decoder, which utilizes self-attention mechanisms to capture complex information within the semantic representation and enables high-precision prediction:
\begin{equation}
  \begin{aligned}
    \widetilde{FMQS}_{IFEM}= &\phi_{IFEM}(\mathcal{I}_{IFEM}), \\
   \widetilde{FMQS}_{BFEM}= &\phi_{BFEM}(\mathcal{I}_{BFEM}).
  \end{aligned}
\end{equation}

During training, the Mean Squared Error (MSE) is adopted as the regression objective. The regression network is optimized by minimizing the squared difference between the predicted score \(\widetilde{FMQS}\) and the truth score FMQS:
\begin{equation}
    \begin{aligned}
        \mathcal{L}{\text{reg}} = \frac{1}{S} \sum{i=1}^S\left( \widetilde{FMQS}^{(i)} - FMQS^{(i)} \right)^2,
    \end{aligned}
\end{equation}
where \( FMQS^{(i)} \) denotes the truth score for the \( i-th \) sample, obtained from the previously defined scoring framework. This design enables efficient prediction of feature quality scores from semantically aligned representations, providing reliable quantitative support for evaluating the training maturity of functional modules.
\subsection{Functional Module Evaluation}
In models such as BEVDepth \cite{li2023bevdepth}, BEVFormerV2 \cite{yang2023bevformer}, and CLIP-BEVFormer\cite{pan2024clip}, introducing auxiliary losses has significantly improved detection performance by enabling multi-task learning and enhancing feature map quality. Inspired by these advances and to validate the effectiveness of our proposed evaluation method in optimizing BEVFormer, we integrate the trained CLIP-based Feature Map Quality Evaluation Network (CLIP-FMQE-Net) into the BEVFormer training pipeline with frozen parameters.

Specifically, the feature map-truth encoder projects feature maps and their corresponding ground truth into a unified semantic space. The quality score prediction head then generates the Feature Map Quality Score (FMQS), which reflects both model performance and semantic consistency. This score is incorporated into training as the following auxiliary loss:
\begin{equation}
    \begin{aligned} 
        \mathcal{L}_{FMQS}^{IFEM} = 1-\widetilde{FMQS}_{IFEM} , \\
        \mathcal{L}_{FMQS}^{BFEM} = 1-\widetilde{FMQS}_{BFEM}, 
    \end{aligned}
\end{equation}
where \(\mathcal{L}_{FMQS}^{IFEM}\) and \(\mathcal{L}_{FMQS}^{BFEM}\) denote the FMQS losses for the Image Feature Extraction Module and the BEV Feature Extraction Module, respectively, with \(\widetilde{FMQS}_{IFEM}\) and \(\widetilde{FMQS}_{BFEM}\) being the corresponding FMQS.

The total loss is formulated as a weighted combination of the original BEV detection loss \(\mathcal{L}_{BEV}\) and the FMQS losses \(\mathcal{L}_{FMQS}^{IFEM}\) and \(\mathcal{L}_{FMQS}^{BFEM}\) from the IFEM and BFEM modules, jointly optimizing task performance and feature quality to improve the model's generalization capability:
\begin{equation}
    \begin{aligned}
        \mathcal{L}_{total} =\;& w_{BEV} \cdot \mathcal{L}_{BEV} 
        + w_{IFEM} \cdot \mathcal{L}_{FMQS}^{IFEM}\\
        &+ w_{BFEM} \cdot \mathcal{L}_{FMQS}^{BFEM}, 
    \end{aligned}
\end{equation}
where \(w_{BEV}\), \(w_{IFEM}\), and \(w_{BFEM}\) are the weighting coefficients for each loss component.
\section{Experiment}
This work follows the experimental setup from \cite{zhang2024unveiling}, ensuring consistency in dataset selection, evaluation metrics, and multi-module training configurations for BEVFormer.
\subsection{Computation of FMQS}

We conduct training using a multi-module learning approach, with eight algorithmic configurations for BEVFormer, each evaluated across eight training stages (as shown in Table~\ref{tab 1}). The configuration–stage combination with the highest NDS is selected as the SOTA model. Specifically, the sixth configuration, VoV-SCA-RCF, achieves the highest NDS of 0.60 at Stage 7. The heatmap in Figure~\ref{fig2}(a) illustrates the performance distribution across combinations, confirming the performance improvements of the deeper training stages and the use of the VoVNet backbone.

Building on this foundation, we construct a unified FMQS evaluation system by comparing all configuration–stage combinations with the SOTA model at both the macro model level and micro feature map level.

At the macro level, the NDS of the SOTA model serves as the reference, with its model task granularity score $Score_{Model}$, set to 1. A higher NDS ensures that each functional module effectively extracts upstream features and generates high-quality feature maps. The bar chart clearly illustrates differences among configuration–stage combinations.

At the micro level, we use CS-CosSim to measure the similarity between feature maps and those of the SOTA model in both the channel and spatial dimensions, reflecting semantic consistency.

Together, the multi-granularity evaluation method provides a comprehensive measure of feature map quality and offers a quantitative basis for further model optimization.

\begin{table}[t]
\centering
\setlength{\tabcolsep}{2.8pt}
\small
\begin{tabular}{lcccccccc}
\hline
\textbf{Module Algorithms} & \multicolumn{8}{c}{\textbf{Training Stage}} \\
\cline{2-9}
\textbf{Configuration} & \textbf{1} & \textbf{2} & \textbf{3} & \textbf{4} & \textbf{5} & \textbf{6} & \textbf{7} & \textbf{8} \\
\hline
\textbf{Res50-SCA-TSA} & 0.23 & 0.28 & 0.31 & 0.35 & 0.38 & 0.40 & 0.42 & 0.44 \\
\textbf{Res50-SCA-RCF} & 0.28 & 0.32 & 0.35 & 0.38 & 0.42 & 0.44 & 0.46 & 0.48 \\
\textbf{Res50-GKT-TSA} & 0.22 & 0.27 & 0.31 & 0.35 & 0.37 & 0.40 & 0.42 & 0.43 \\
\textbf{Res50-GKT-RCF} & 0.23 & 0.28 & 0.33 & 0.36 & 0.39 & 0.42 & 0.45 & 0.46 \\
\textbf{VoV-SCA-TSA}  & 0.39 & 0.42 & 0.47 & 0.49 & 0.52 & 0.54 & 0.55 & 0.55 \\
\textbf{VoV-SCA-RCF}  & 0.42 & 0.47 & 0.53 & 0.56 & 0.58 & 0.58 & 0.60 & 0.59 \\
\textbf{VoV-GKT-TSA}  & 0.38 & 0.42 & 0.46 & 0.49 & 0.52 & 0.55 & 0.55 & 0.56 \\
\textbf{VoV-GKT-RCF}  & 0.41 & 0.45 & 0.50 & 0.53 & 0.58 & 0.58 & 0.59 & 0.59 \\
\hline
\end{tabular}
\caption{BEVFormer MML: NDS Results on 3D Object Detection}
\label{tab 1}
\end{table}

\subsection{Training Settings for FMQS Prediction}
We construct a regression dataset named FMQS-Dataset based on the nuScenes-mini dataset, which contains 404 keyframe samples. The dataset is split into a training set (323 samples) and a test set (81 samples) with a 4:1 ratio. For all 8 configurations and 8 training stages in the multi-module training setup, we retain the model weights at each stage and store the output feature maps from both the IFEM and BFEM modules. The corresponding FMQS are used as regression labels.
\begin{table}[t]
    \centering
    \setlength{\tabcolsep}{2.8pt} 
    \small
    \begin{tabular}{lcccc}
    \hline
    \textbf{Module Configuration} & \textbf{MSE} & \textbf{MAE} & \textbf{$R^2$} & \textbf{MAPE(\%)} \\
    \hline
    Res50-SCA-TSA & 0.0039 & 0.0478 & 0.8690 & 7.1825 \\
    Res50-SCA-RCF & 0.0041 & 0.0487 & 0.8517 & 7.3432 \\
    Res50-GKT-TSA & 0.0042 & 0.0496 & 0.8462 & 7.5199 \\
    Res50-GKT-RCF & 0.0042 & 0.0499 & 0.8472 & 7.4694 \\
    VoV-SCA-TSA   & 0.0039 & 0.0490 & 0.8565 & 7.4331 \\
    VoV-SCA-RCF   & 0.0038 & 0.0481 & 0.8564 & 7.2764 \\
    VoV-GKT-TSA   & 0.0041 & 0.0482 & 0.8573 & 7.6509 \\
    VoV-GKT-RCF   & 0.0043 & 0.0498 & 0.8563 & 7.4435 \\
    \hline
    \textbf{Average} & \textbf{0.0041} & \textbf{0.0489} & \textbf{0.8551} & \textbf{7.4149} \\
    \hline
    \end{tabular}
    \caption{FMQS Regression Results for the IFEM Module}
    \label{tab 2}
\end{table}

The ground truth encoder adopts the CLIP-ViT-B/32 text encoder from HuggingFace, pretrained on 400 million image–text pairs through contrastive learning. This pretrained model provides strong cross-modal semantic alignment capabilities and improves transfer learning, enabling faster convergence and higher prediction accuracy, especially under limited data conditions \cite{doroudian2024clip}.

The feature map encoder adopts a four-layer convolutional neural network (CNN).

For IFEM, the input \(\mathcal{F}_{img} \in \mathbb{R}^{6,256,15,25}\) is reshaped to \(\mathbb{R}^{6\times256,15,25}\), then passed through  \(3 \times 5\) convolutional kernels, max pooling, ReLU activation, and a fully connected layer $(6 \times 512)$ to project the output into a unified semantic space. Ground-truth texts are encoded in a six-camera format to align with image features.

For BFEM, the input \(\mathcal{F}_{bev} \in \mathbb{R}^{256,50,50}\) goes through four convolutional layers with \(3 \times 3\) kernels, each followed by \(2 \times 2\) max pooling and ReLU activvation. The final output is mapped to a 512-dimensional semantic vector for FMQS prediction.

During training, we jointly train the feature map–ground truth encoder and the FMQS prediction head under the CLIP-FMQE-Net framework, with separate models for IFEM and BFEM. The model is trained for 20 epochs with a batch size of 64 and an initial learning rate of $10^{-4}$, which is adjusted via cosine annealing. All experiments are conducted on an NVIDIA Tesla A100 GPU with 40GB memory.
\begin{figure*}[t]
  \centering
\includegraphics[width=1.5\columnwidth]{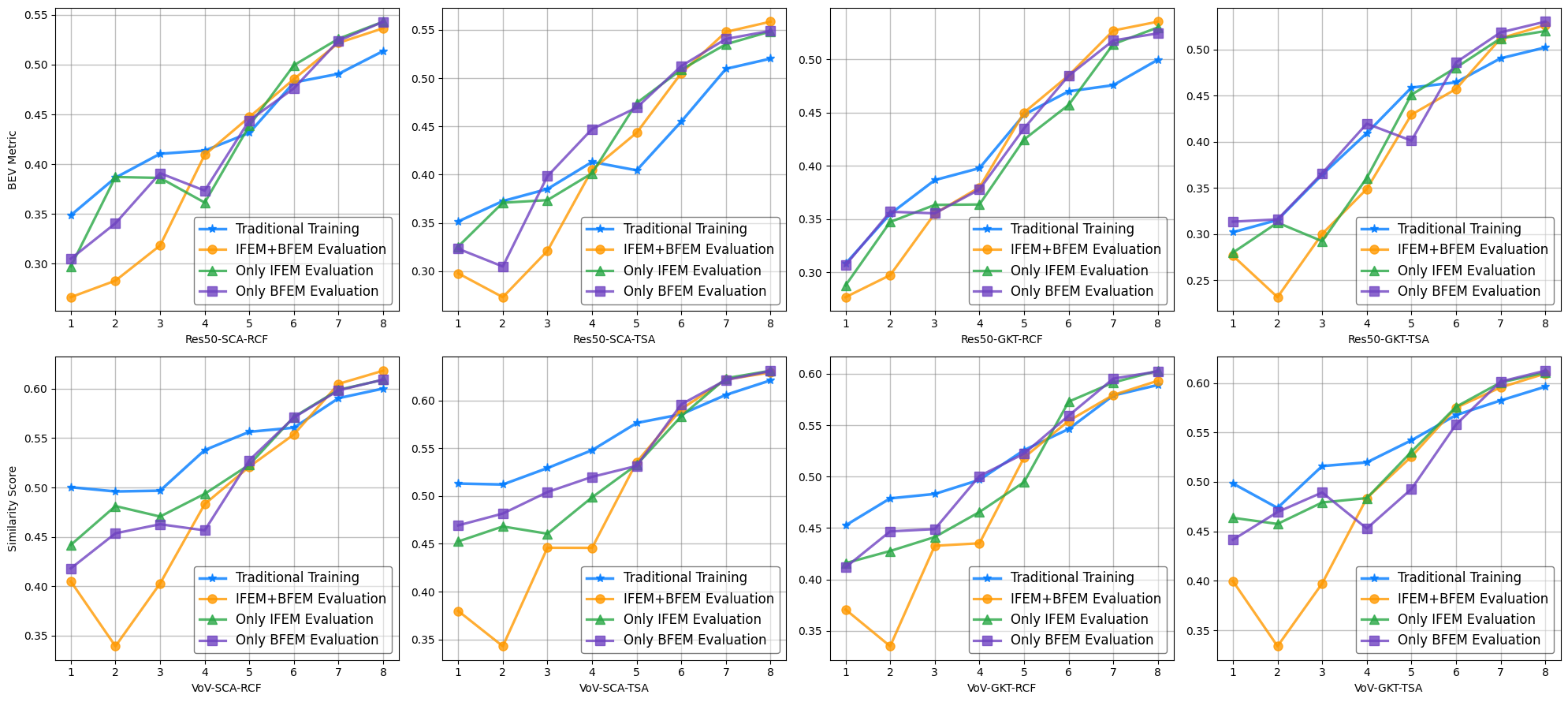}
\caption{Training Results of BEVFormer with Integrated Independent Evaluation Modules for 3D Obstacle Detection}
\label{fig3}
\end{figure*}
\subsection{FMQS Regression Results and Analysis}
To evaluate the prediction performance of CLIP-FMQE-Net, we adopt four standard regression metrics: Mean Squared Error (MSE), Mean Absolute Error (MAE), Coefficient of Determination \(R^2\), and Mean Absolute Percentage Error (MAPE), defined as:
\begin{equation}
\text{MSE} = \frac{1}{n} \sum_{i=1}^{n} (y_i - \hat{y}_i)^2
\end{equation}

\begin{equation}
\text{MAE} = \frac{1}{n} \sum_{i=1}^{n} |y_i - \hat{y}_i|
\end{equation}

\begin{equation}
R^2 = 1 - \frac{\sum_{i=1}^{n} (y_i - \hat{y}_i)^2}{\sum_{i=1}^{n} (y_i - \bar{y})^2}
\end{equation}

\begin{equation}
\text{MAPE} = \frac{1}{n} \sum_{i=1}^{n} \left| \frac{y_i - \hat{y}_i}{y_i} \right| \times 100\%
\end{equation}

where \(n\) is the number of samples,\(y_i\) is the ground truth FMQS, \(\hat{y}_i\) is the predicted score, and \(\bar{y}\) is the mean of the ground truth FMQS.

These metrics jointly evaluate the model from the perspectives of error magnitude, stability, and interpretability.

Table~\ref{tab 2} reports the IFEM regression results across eight configurations. VoV-SCA-RCF achieves the lowest MSE (0.0038), MAE (0.0481), and MAPE (7.28\%), while Res50-SCA-TSA attains the highest \(R^2\) (0.869), showing superior fitting. The average performance across all configurations (MSE: 0.0041, MAE: 0.0489, \(R^2\): 0.8551, MAPE: 7.41\%) reflects high prediction accuracy and robustness.

Table~\ref{tab 3} summarizes the BFEM results. Res50-SCA-TSA yields the lowest MSE (0.0050) and MAE (0.0544), while Res50-GKT-RCF records the highest \(R^2\) (0.8192). Although VoV-GKT-TSA shows the highest MAE (0.0616), all configurations remain within a reasonable error range. The average values (MSE: 0.0055, MAE: 0.0584, \(R^2\): 0.7936, MAPE: 8.86\%)  indicate high accuracy and strong robustness.
\begin{table}[t]
\centering
\small
\setlength{\tabcolsep}{2.8pt} 
\begin{tabular}{lcccc}
\hline
\textbf{Training Configurations} & \textbf{MSE} & \textbf{MAE} & \textbf{$R^2$} & \textbf{MAPE (\%)} \\
\hline
\textbf{Res50-SCA-TSA}   & 0.0050 & 0.0544 & 0.7899 & 8.3755 \\
\textbf{Res50-SCA-RCF}   & 0.0058 & 0.0591 & 0.7869 & 8.9624 \\
\textbf{Res50-GKT-TSA}   & 0.0057 & 0.0579 & 0.8085 & 8.5452 \\
\textbf{Res50-GKT-RCF}   & 0.0055 & 0.0573 & 0.8192 & 8.6435 \\
\textbf{VoV-SCA-TSA}     & 0.0064 & 0.0600 & 0.7657 & 8.9619 \\
\textbf{VoV-SCA-RCF}     & 0.0058 & 0.0599 & 0.7967 & 9.1983 \\
\textbf{VoV-GKT-TSA}     & 0.0065 & 0.0616 & 0.7764 & 9.3700 \\
\textbf{VoV-GKT-RCF}     & 0.0055 & 0.0571 & 0.8052 & 8.8331 \\
\hline
\textbf{Average}         & \textbf{0.0055} & \textbf{0.0584} & \textbf{0.7936} & \textbf{8.8612} \\
\hline
\end{tabular}
\caption{FMQS Regression Results for the BFEM Module}
\label{tab 3}
\end{table}

\begin{table}[t]
\centering
\small
\setlength{\tabcolsep}{6.0pt} 
\begin{tabular}{lccc}
\hline
\textbf{Training} & \textbf{IEFM} & \textbf{BEFM} &\textbf{IEFM +} \\
\textbf{Combination} &\textbf{(\%)} &\textbf{(\%)} &\textbf{BEFM (\%)} \\
\hline
\textbf{Res50-SCA-TSA}  & 5.72 & 5.74 & 4.46 \\
\textbf{Res50-SCA-RCF}  & 5.54 & 5.42 & 7.36 \\
\textbf{Res50-GKT-TSA}  & 4.94 & 6.07 & 7.15 \\
\textbf{Res50-GKT-RCF}  & 5.58 & 3.55 & 4.84 \\
\textbf{VoV-SCA-TSA}    & 1.52 & 1.48 & 3.02 \\
\textbf{VoV-SCA-RCF}    & 1.58 & 1.63 & 1.32 \\
\textbf{VoV-GKT-TSA}    & 2.24 & 2.36 & 0.70 \\
\textbf{VoV-GKT-RCF}    & 2.72 & 2.40 & 2.23 \\
\hline
\textbf{Average}        & \textbf{3.73} & \textbf{3.58} & \textbf{3.89} \\
\hline
\end{tabular}
\caption{NDS Improvement (\%) from Functional Module Integration in BEVFormer.}
\label{tab 4}
\end{table}
Comparing the two modules, IFEM demonstrates better regression performance with lower MSE and higher \(R^2\), indicating stronger fitting and lower prediction error. Both modules, however, achieve high accuracy and robustness (MSE $<$ 0.006, \(R^2\) $>$ 0.76), validating the effectiveness of the cascaded architecture composed of the feature map–ground truth encoder and the FMQS prediction head. This provides a reliable quantitative benchmark for optimizing BEVFormer.
\subsection{Evaluation of Training with Integrated Functional Module Assessment}

Tables~\ref{tab 2} and~\ref{tab 3} validate the regression performance of CLIP-FMQE-Net in predicting FMQS when integrated into BEVFormer training. By introducing FMQS as an auxiliary loss, we demonstrate the effectiveness of the proposed dual-granularity dynamic weighted scoring system, which contributes to improved model performance.

Figure~\ref{fig3} compares different training strategies. The orange curve—joint integration of image and BEV extraction evaluation modules—consistently achieves higher NDS than traditional training (blue) and single-module integration (green/purple), confirming its advantage in guiding feature learning.

Table~\ref{tab 4} further quantifies this improvement. Under the Res50-SCA-TSA configuration, image and BEV modules individually improve NDS by 5.72\% and 5.74\%. Their combined integration yields up to 7.36\% gain (Res50-SCA-RCF), demonstrating the benefits of multi-module collaborative optimization.

FMQS effectively reflects the influence of feature map quality on model performance, supporting comprehensive evaluation and improving NDS and other detection metrics. The combined IFEM+BFEM modules offer rich supervision that enhances training dynamics. Leveraging CLIP’s cross-modal alignment, the framework provides diverse quality-aware signals for complementary visual and semantic analysis, facilitating better feature learning and metric optimization.
\section{Conclusion}
This study presents an independent evaluation method based on FMQS, enabling fine-grained assessment and optimization within end-to-end training. We introduce a dual-granularity dynamic weighted scoring system and propose CLIP-FMQE-Net, a CLIP-based feature map quality evaluation network comprising a feature map encoder, a text encoder, and a prediction head. This architecture aligns feature maps and ground truth text in a shared semantic space and accurately predicts FMQS.

Experiments show that CLIP-FMQE-Net achieves high prediction accuracy and robustness. When integrated into BEVFormer, it improves 3D object detection on nuScenes, achieving an average NDS gain of 3.89\%. Joint use of image and BEV feature evaluation modules further enhances performance, validating the effectiveness of FMQS and the proposed evaluation framework.

Leveraging CLIP's cross-modal alignment, this approach provides a principled paradigm for training autonomous driving perception models. Future work will explore using FMQS as an auxiliary loss and extending it to broader tasks and architectures to evaluate its generalizability.
\section*{Acknowledgments}
This work was supported by the National Key R\&D Program of China, Project "Development of Large Model Technology and Scenario Library Construction for Autonomous Driving Data Closed-Loop" (Grant No.2024YFB2505501), the Guangxi Science and Technology Major Program, Project "Research and Industrialization of Cost-Effective Urban Pilot Driving Key Technologies" (Grant No.AA24206054), and the Independent Research Project of the State Key Laboratory of Intelligent Green Vehicle and Mobility, Tsinghua University (No.ZZ-GG-20250405).

\bibliographystyle{plain}  
\bibliography{arxiv}

\vfill
\end{document}